\documentclass[letterpaper]{article} 
\usepackage[preprint]{aaai2027}  
\usepackage[hyphens]{url}  
\usepackage{graphicx} 
\usepackage{natbib}  
\usepackage{caption} 
\usepackage{booktabs}
\usepackage{amsmath}
\newcommand{\rtilde}{\ensuremath{\tilde{r}}}
\title{Train the Model, Not the Reader: Decodability Supervision for Verifiable
Activation Explanations}
\author{
    Hiskias Dingeto
}
\affiliations{
    StackOne Technologies\\
    hiskias@stackone.com
}

\begin{document}

\maketitle

\begin{abstract}
Natural-language autoencoders score explanations of hidden activations by
reconstruction. An explanation is deemed faithful if the activation can be
regenerated from it. The test is structurally insensitive to individual false
claims. Flipping one to a false alternative often leaves the reconstruction unchanged, so it is
never penalized. We show the test can be passed without faithfulness, in two distinct ways. On a
released Qwen-2.5-7B verbalizer, explanations reconstruct well above chance while
as few as ${\sim}2\%$ of their specific claims are ones the reconstruction depends on. The score therefore
tracks the input's gist, not its specific facts. Under exact synthetic
ground truth, standard training consistently develops co-adapted private codes, false
wording the reconstruction depends on. Fixes that leave the
target model unchanged do not help. We contribute two audit protocols (the
comparison of grounding and truth, and the swap to an independent evaluator) and RECAP (Readable Encodings via
Co-trained Auxiliary Predictors), linear heads trained alongside the target model
to keep designated content decodable. On RECAP-trained sandbox models, fresh
verbalizers state the designated content truly and the codes vanish, at a
$+0.001$-nat cost. This replicates on a pretrained Pythia-160M under a rule for designing targets we identify. The content becomes reliably decodable by a probe, though a fresh
verbalizer conveys it only in part (truth 0.44--0.46 against a near-zero
control). For interpretability, the audit sets a claim-level standard of evidence
for activation explanations. For AI safety, RECAP makes designated content
checkable against probes rather than taken on trust from verbalizer prose a model can game. On the pretrained model, an independent probe reliably scores the
verbalizer's true claims above its false ones (AUC 0.96, versus 0.82 without
RECAP). Checking the probe therefore catches false claims about designated content. When an adversary edits an
explanation to maximize the score while lying, it suppresses about
87\% of its lie penalty. Yet the
RECAP probe still flags the lies at AUC 0.95, while the control probe collapses to
chance at 0.51.
\end{abstract}

\section{Introduction}

Natural-language autoencoders (NLAs) let a model explain its own hidden
states: a verbalizer turns an activation into a text explanation, a
reconstructor turns the explanation back into an activation, and the quality
of the round trip is treated as a test of each explanation's faithfulness
\citep{nla2026}. The idea is attractive for oversight because it is
unsupervised and self-checking, with no labels or human in the loop, just a
number attached to every explanation. Released
verbalizer/reconstructor pairs exist for several open models, and a growing
family of methods trains activation readers with related objectives
\citep{latentqa2024,activationoracles2025,pcd2025}.

This reconstruction test has a structural gap. Reconstruction only checks that the
explanation contains enough to regenerate the activation. It does not
penalize false additions, because the objective has no preference among
claim values that induce the same reconstruction. This gap is a property of the objective
itself, present regardless of model or scale.

We show that in practice the test is passed in two ways we identify, neither
of which is faithful at the level of individual claims. By \emph{gist}, on a released
Qwen-2.5-7B-based NLA, explanations reconstruct well above chance. Yet
our audit, which flips each claim to test whether the reconstruction depends on it, detects grounding in only ${\sim}2\%$ of their salient specific
claims. This ${\sim}2\%$ is a lower bound, since the flip test has low sensitivity and likely undercounts true grounding. By \emph{codes}, in a synthetic
sandbox with exact ground truth, co-trained verbalizer/reconstructor pairs
consistently develop private codes, false wording that the reconstruction depends
on (gaps between grounded and true rates of 0.30--0.80),
using standard training across two scenarios. The two natural training fixes
also fail. One minimizes the distance between a hidden state and its reconstruction,
which removes content a probe could previously read. The other rewards the verbalizer for grounding,
which raises the score by changing which sentences the metric counts, not by grounding more claims.

\begin{figure*}[t]
\centering
\includegraphics[width=\textwidth]{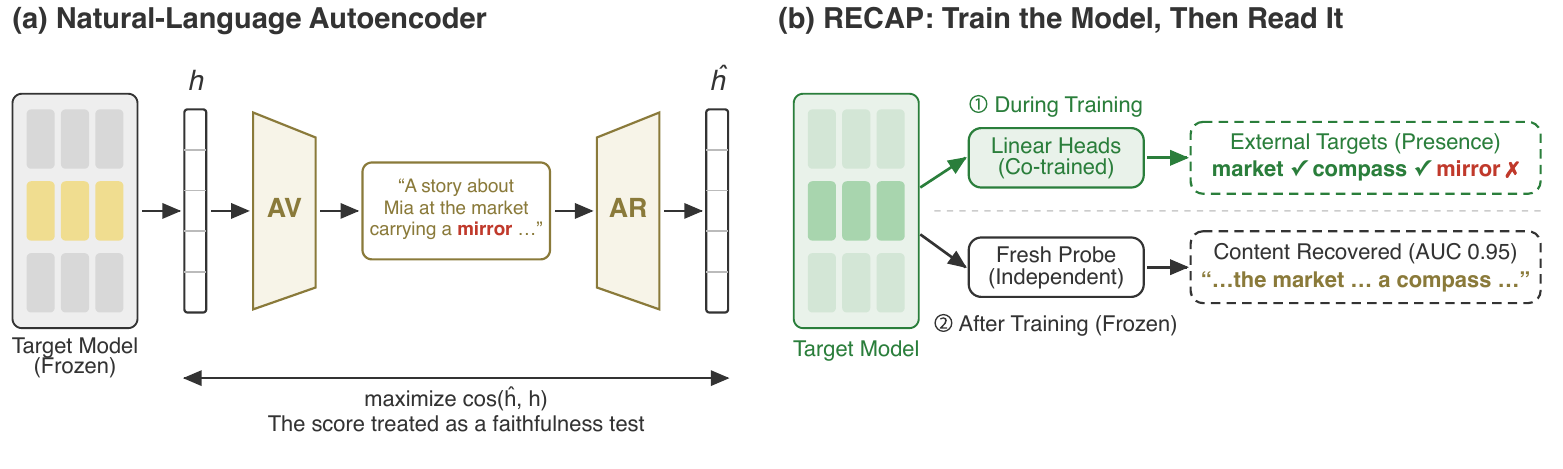}
\caption{The two architectures. (a) A natural-language autoencoder (prior work)
\citep{nla2026}: AV maps an activation to text, AR maps it back, and
reconstruction quality is the test; only AV and AR train. The score can stay
high while the text asserts an ungrounded specific (red ``mirror''). (b) RECAP
(this work) trains the \emph{target model}. Top: co-trained linear heads keep
designated content decodable against external targets. Bottom: after training,
an \emph{independent} fresh probe (not the heads) recovers the content (AUC
0.95) and a fresh verbalizer states it truly.}
\label{fig:arch}
\end{figure*}

The intervention that survives our controls trains the \emph{target
model}, not the activation reader (Figure~\ref{fig:arch}b): RECAP, linear heads on
external targets co-trained with the model (\S\ref{sec:recap}). Fresh NLAs
then state the designated content faithfully in the sandbox under both co-trained and
independent evaluators, and the codes are absent. Both
alternatives motivated by interpretability fail for identified reasons.
The reconstruction objective collapses decodable content, and the approach using a frozen probe goes stale
under representation drift. A realignment that needs no labels fully reverses this.

Our contributions:
\begin{itemize}
\setlength{\itemsep}{0pt}
\setlength{\parskip}{0pt}
\item \textbf{Claim-level audit.} Flipping each salient claim of explanations scored by reconstruction, with validity controls for each edit, shows that a released verbalizer passes the test even though its specific claims are largely ungrounded; the score survives paraphrase but collapses when content words are masked, so gist carries it, not specific wording.
\item \textbf{Private codes.} Co-trained verbalizer/reconstructor pairs develop private codes in a sandbox with exact ground truth; the comparison of grounding and truth and the swap to an independent evaluator detect and quantify them.
\item \textbf{RECAP.} Where a reconstruction objective collapses decodable content and a frozen probe goes stale under drift, RECAP instead trains the target model. This restores claim-level faithfulness in the sandbox and transfers to Pythia-160M, on which designated content becomes decodable by a probe at a measured cost.
\item \textbf{Adversarial monitor.} An adversary that edits the explanation to game the deployed reconstruction score suppresses ${\sim}87\%$ of its lie penalty on both models, yet RECAP's activation probe still flags the lies (AUC $0.95$) while the control probe collapses to chance ($0.51$). RECAP's decodability thus survives text gaming, and the reconstruction score must not be used as a truth signal.
\end{itemize}

\section{Preliminaries and Experimental Setup}
\label{sec:prelim}

\subsection{The Reconstruction Test and Audit Protocols}

A natural-language autoencoder (NLA) pairs an \emph{activation verbalizer}
(AV), which generates a text explanation $z$ of a hidden activation $h$, with
an \emph{activation reconstructor} (AR), which maps $z$ back to an estimate
$\hat{h}$. The AV is trained against reconstruction
reward; with target model weights $\theta$ frozen and $h$ its tap
activation, the pair optimizes
\begin{equation}
\mathcal{J}_{\mathrm{read}} \;=\;
\max_{\mathrm{AV},\,\mathrm{AR}}\;
\cos\!\big(\mathrm{AR}(\mathrm{AV}(h)),\, h\big),
\label{eq:jread}
\end{equation}
and the reconstruction quality of a produced explanation is treated as its
faithfulness test (Figure~\ref{fig:arch}a). \citet{nla2026} are explicit about this. They present rising
reconstruction as tracking explanation informativeness, and warn that
nothing in the objective forces faithfulness. But that
score is still the number these NLAs report, and the main criterion their users apply. We ask what that score indicates about the
individual claims of each explanation. We measure
reconstruction with a centered cosine normalized to the floor. Let $\mathrm{floor}$ be the mean cosine between
reconstructions and \emph{mismatched} activations (chance level). Then
$\rtilde = (\cos(\hat{h}, h) - \mathrm{floor}) / (1 - \mathrm{floor})$:
0 at chance, 1 at perfect reconstruction. For the released NLA we use the Qwen-2.5-7B layer-20 AV/AR pair~\citep{qwen2025}.

\emph{Grounding.} For a claim $c$ in $z$, let $z_{\neg c}$ be $z$ with $c$
alone minimally flipped (a minimal-pair counterfactual edit, which we call a flip). The claim's grounding is the reconstruction drop
\begin{equation}
\Delta\rtilde(c) = \rtilde(z) - \rtilde(z_{\neg c}) ,
\end{equation}
and $c$ is \emph{grounded} when $\Delta\rtilde(c) \geq \tau$ for a threshold $\tau$.
Grounding here is the sensitivity of the \emph{reconstruction} to a claim flip.
Our thesis is that this is not the same as the model's own sensitivity to the flip, and that the two come apart in practice. We check each edit for validity. An edit that changes more than its target claim can inflate apparent grounding severalfold (Appendix~A). Crossing
grounding with truth gives four cases. A
claim is grounded or not, and true of the source or not. We
focus on the case where wording is \emph{grounded but false}. Here the
reconstruction depends on the wording, although its stated content is false. When
such wording is a convention specific to the co-trained pair (shown by the
swap to an independent evaluator, \S\ref{sec:codes}), we call it a \emph{co-adapted private code}. These
labels are operational; no intentional stance is implied. Let $\rtilde_{\mathrm{co}}$ and
$\rtilde_{\mathrm{ind}}$ be the scores under the co-trained and an independent
reconstructor. The score's \emph{evaluator gap} is then
$\rtilde_{\mathrm{co}} - \rtilde_{\mathrm{ind}}$.

\emph{Terms.} The target model's activation is read at a chosen
layer and position, the \emph{tap}. An activation can be read by the reconstructor AR or by a \emph{probe}, a freshly fitted
linear decoder \citep{alain2017probes,belinkov2022probing}. \emph{Gist} is
the input's overall meaning, as opposed to its specific facts. Content is
\emph{designated} when an auxiliary predictor is co-trained to predict it
(\S\ref{sec:recap}). Content is \emph{decodable}
when a freshly fitted linear probe can read it. Faithfulness in this paper is always claim-level. We judge each individual claim in an
explanation as true of the activation and grounded in it, rather than judging the explanation as a whole. We do not use
faithfulness to mean whether explanations describe the model's downstream computation, or grounding to mean grounding in the world.

\paragraph{Settings.}\label{sec:setup} We evaluate in three settings. The first is a released
Qwen-2.5-7B layer-20 NLA on in-distribution web text ($n{=}1{,}517$ audited
claims). The second is a \emph{sandbox} of two synthetic scenarios, a \emph{story} and a \emph{marketplace},
each built by filling the slots of a fixed template, so that the true content of every activation is known exactly and computable without a judge; here the tap position fixes what the activation can hold (retained, faded, or unread slots). The third is
continued pretraining of Pythia-160M~\citep{biderman2023pythia} with self-supervised heads, where \emph{tax}
is held-out language-modeling loss minus a shared control. Full protocols, scenario
grammars, seeds, and hyperparameters are in Appendices~D and~K.

\section{Auditing Reconstruction-Scored Explanations}
\label{sec:passed}

The two ways of passing the test have different mechanisms. On the released NLA the missing specifics
are largely \emph{absent} from the tap, so the high reconstruction score depends on gist rather than
on specific content in the activation. In the sandbox the content is present, but
the verbalizer states it in wording that only its co-trained reconstructor understands. The swap to an independent evaluator exposes this. Only the
sandbox failure is a pure artifact of the metric, where the content is there but the score is fooled. The released NLA's failure instead reflects content genuinely missing from the activation.

\subsection{Counterfactual Audit of a Released Verbalizer}

On in-distribution web text, the released NLA's explanations reconstruct
at $\rtilde = 0.84$, yet the audit finds grounding in only about $2\%$ of
the specific claims an LLM finds salient ($4.2\%$ / $2.1\%$ /
$1.6\%$ at $\tau = 0.02$ / $0.05$ / $0.10$; full table in Appendix~E).
Two qualifications apply. First, this is a lower bound, since reconstruction has low
sensitivity and undercounts true grounding; a control that appends and re-ranks
recovers a topic known to be encoded at only $28\%$ raw. Second, the more robust statistic is the relative gap between gist and specifics. A probe recovers specific facts about $3\times$ less accurately than gist (${\sim}6\%$
versus ${\sim}18\%$ chance-corrected at each position; mean-pooled gist reaches
${\sim}50\%$; Appendix~E).

Overall, the NLA reconstructs well despite depending weakly on most of the $n{=}1{,}517$ audited specific claims. A transform control confirms this directly. Reconstruction survives
paraphrase but collapses when content words are masked (Appendix~E), so gist, not specific wording, carries the score. The effect holds across layers 16--27 and is specific to readouts in the reconstruction style. The model's own output head does see these late-layer specifics in its next-token predictions (Appendix~E). Appendix~E also reports bootstrap intervals and a noise floor, estimated from flips that carry no grounding signal, sitting below the smallest threshold.

One could also train the verbalizer toward grounding directly. We tried
two approaches: rejection sampling and GRPO against a grounding reward scored at the clause level. Both improved their proxy while the independent claim-level audit
stayed null at every $\tau$ (confidence intervals straddle zero). Text
statistics attribute the gains to sentence restructuring, merging and
packaging the units the policy controls, not to increased claim-level
grounding. This motivates the sandbox, a setting where every
quantity is exact.

\subsection{Audit Under Synthetic Ground Truth}
\label{sec:codes}

A second way of passing the test appears under synthetic data whose true content is known exactly. In
every sandbox run the toy NLA asserts every slot confidently while passing the
test at $\rtilde$ 0.71--0.86 (both scenarios; representative example in
Appendix~D). Truth tracks decodability only loosely. The
fully decodable slot is stated truly at 0.44--0.96 across seeds (yet grounded
0.76--0.88 in all), partially decodable slots at 0.15--0.26, and unread slots are stated truly no better than
chance. The test does not distinguish truthful
claims from false ones, because the training signal carries no such requirement.

Beyond truth rates, the co-trained pair produces a second systematic phenomenon. In 5/5 runs of standard training it develops \emph{co-adapted private codes}:
wording choices on which the reconstruction depends even though the wording is false, with
gaps between grounded and true rates of 0.30--0.80. The comparison of grounding and truth is
the detector. A code is grounded \emph{and} false. Codes can occupy several slots at once. In the marketplace scenario one code occupies an
\emph{unread} slot, since any wording the reconstructor relies on suffices,
whether or not its content is true or even possible. In that case the pair is not describing
the activation but using the sentence as a communication channel.

The swap to an independent evaluator provides evidence that the channel is a convention
specific to the co-trained pair. Under that evaluator (trained without exposure to the audited
pair's outputs), grounding on slots carrying codes collapses from 0.53--0.89 to
0.08--0.23, while grounding of true claims survives. The swap locates every
claim in the table cross-tabulating grounding and truth. The cross-evaluator
$\rtilde$ gap (0.17--0.38 on standard runs versus ${\sim}0.003$ on RECAP
runs, \S\ref{sec:recap}) is the evaluator gap of the passing score. This gap reflects
co-adaptation, not a weaker independent evaluator. If the evaluator were simply worse, grounding of true claims would drop too; instead it survives. Independence holds at the phrasing
level too. An independent evaluator trained on a \emph{disjoint} paraphrase
family (different frames and slot order, sharing only the slot words)
reproduces the same pattern within ${\sim}0.05$ across three seeds. Grounding
on slots carrying codes stays collapsed while genuine grounding survives (RECAP slots at 1.00).
The channel is therefore private to the co-trained pair, not a convention shared across templates.

Finally, verbalizers from standard training \emph{under-read} even content that is
present. In the marketplace scenario, slots that are 55--72\% decodable from the target
model are stated truly only 15--26\% of the time. This is a \emph{readout gap} (content
present but not conveyed by the verbalizer), in
addition to the \emph{storage gap} (content absent from the activation). The intervention below also closes it.

\section{Representation Collapse and Probe Drift}
\label{sec:failure}

We now explain why the two natural fixes fail, through two mechanisms. The first failure mode is representation collapse. The most natural objective on the target side moves activations toward their
own reconstruction. The probe's recovery and the model's computation then agree. This objective collapses decodable content. At every tested
strength, regardless of initialization, it removes decodable slot
content while perplexity is unchanged. Freshly retrained probes lose accuracy too, so the
information itself is gone, not merely rotated into a subspace probes cannot read. The mechanism is consistent
with prior analyses of collapse in non-contrastive self-supervised learning
\citep{simsiam2021,tian2021understanding}.

The reconstruction is a lossy
projection of the state. Increasing the similarity between the state and its reconstruction reduces
variation outside the probe's subspace, so it removes any content the probe
cannot recover.

The second failure mode is representation drift during continued training.
Probes matched to a training model decay to near-chance within 1{,}500
continued training steps and are at chance across seeds (all six
pairs). Yet the information itself remains intact and fully recoverable. An
affine map fitted on paired activations (no labels needed) restores native
probe performance in all nine cases. Decomposing that map (mode sweep
figure in Appendix~H) shows that rotation alone recovers about half the drift;
full recovery needs a non-orthogonal change of rank ${\sim}$16--64. \citet{duan2026monitors} concurrently reports the
same drift and realignment on 2B-parameter and 7B-parameter models.

This finding on probe drift and realignment has a deployment consequence. Probe-based monitoring
of a model that is still training must include an explicit recalibration procedure.
Realignment restores a probe's access to what the target model stores, but
whether the model stores the right content is the underlying problem. Realignment only corrects how the probe reads the representation; RECAP changes what the representation contains. A corollary follows. Because representations differ by an affine map, adapter-based transfer
of probes across models should work \citep{uav2026}.

\section{The RECAP Objective}
\label{sec:recap}

\begin{table}[t]
\centering
\setlength{\tabcolsep}{4pt}
\begin{tabular}{lcc}
\toprule
configuration & tax (nats) & probe AUC \\
\midrule
control (baseline)              & n/a      & 0.77--0.80 \\
$K{=}8$, naive, $w{=}1.0$       & $+0.047$ & 1.000 \\
$K{=}64$, naive, $w{=}1.0$      & $+0.064$ & 0.95--0.97 \\
$K{=}64$, balanced, $w{=}0.1$   & $+0.010$ & 0.97--0.99 \\
$K{=}512$, balanced, $w{=}1.0$  & $+0.14$--$0.20$ & 0.97--0.99 \\
\bottomrule
\end{tabular}
\caption{RECAP at scale: held-out tax (LM loss minus control, nats) and
the AUC of a fresh probe on designated targets. The $K{=}64$ balanced tax is a 3-seed
paired difference (95\% CI $[-0.015, +0.034]$); the other balanced row is a
3-seed range, the rest from a single seed.}
\label{tab:scale}
\end{table}

\subsection{Method}

The approach that survives our controls trains the target model itself, rather than the verbalizer, reconstructor, or probe that reads its activations. RECAP (Readable Encodings via Co-trained Auxiliary Predictors) adds
linear \emph{auxiliary predictors} (heads) during training, so that designated content stays decodable from
selected hidden states. Each head reads the hidden state at the tap layer and predicts an
external target, namely slot values in the sandbox or self-supervised text functions at
scale. We then add the head loss to the language-modeling loss. With head
parameters $\phi$ and external targets $y$,
\begin{equation}
\mathcal{J}_{\mathrm{model}} \;=\;
\min_{\theta,\,\phi}\;
\mathcal{L}_{\mathrm{LM}}(\theta) \;+\;
w\,\mathcal{L}_{\phi}\big(h(\theta),\, y\big),
\label{eq:jmodel}
\end{equation}
Here $w$ weights the head loss against the language-modeling loss. $\mathcal{L}_{\phi}$ is a class-balanced binary cross-entropy for whether each content word is present, plus a cross-entropy for which token it is. Balancing prevents the trivial shortcut of \S\ref{sec:scaling}.

Comparing
Eq.~\ref{eq:jmodel} with Eq.~\ref{eq:jread} states our thesis.
$\mathcal{J}_{\mathrm{read}}$ optimizes the verbalizer and reconstructor against a frozen
model's activations and never touches $\theta$. In contrast,
$\mathcal{J}_{\mathrm{model}}$ trains $\theta$ itself, adding a term to the
language-modeling loss for the ground truth that the verbalizer and reconstructor will need.

RECAP addresses all three failure modes at their source. Drift cannot set in, because the heads co-train with the target model, so no frozen probe is left to go stale. Collapse cannot occur, because the targets are external ground truth, not the model's own reconstruction, so there is no self-referential objective to collapse into. Codes cannot form, because the head reads the activation directly, with no intervening text where a private code could hide.

Because a co-trained predictor could co-adapt
with the model, every decodability number we report comes from
independently fitted probes, never from the heads.
RECAP is decodability supervision in the deep-supervision lineage
\citep{dsn2015,googlenet2015}.

\subsection{Sandbox Results}

On the target model, RECAP produces the intended effect. Designated content reaches 100\%
decodability by \emph{independently fitted} probes, at $+0.001$ nats, in all five runs.
The heads recover only content the activation actually holds. On unread slots, whose values never appear before the tap, they decode nothing in any run. We re-estimate this
tax on pretrained Pythia-160M below, where it is $+0.010$
nats and depends on target design.

What matters is that the effect propagates beyond the heads. Fresh NLAs trained on RECAP target models state the designated
content at ${\sim}1.00$ truth and grounding, under both
the co-trained and the independent evaluator. They flag the remaining
false content at grounding ${\sim}0$. No codes appear in any
RECAP run (emergence ledger in Appendix~D). Both NLAs receive identical
training, including a cold-start SFT on gold descriptions. Only the NLA
trained on activations from the RECAP target model retains near-perfect truth and
grounding, so the difference isolates the effect of target model
decodability.

\begin{figure*}[t]
\centering
\includegraphics[width=0.9\textwidth]{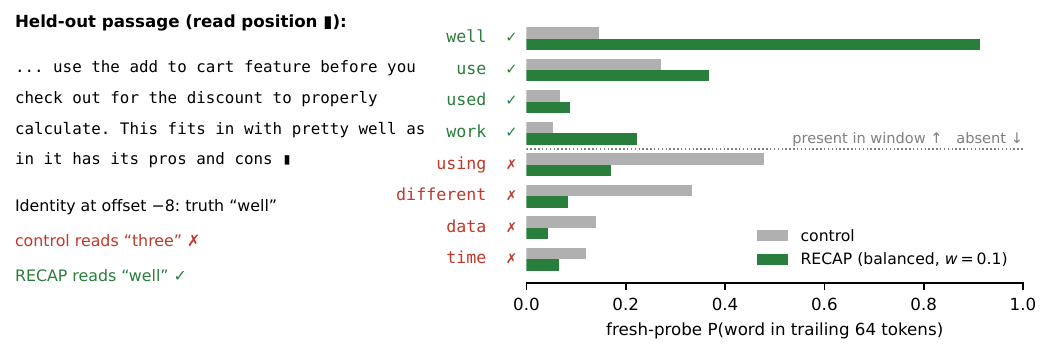}
\caption{Decoding content from a real model's hidden state (one held-out Pile~\citep{gao2020pile}
passage; fresh probes read the layer-6 state at the marked position). On the
RECAP-trained model the probes recover the trailing-window content words and the
offset-8 token; on the control the readout is scrambled. Same probe recipe and
data; only the target model differs.}
\label{fig:probe}
\end{figure*}

\subsection{Scaling to Pythia-160M}
\label{sec:scaling}

We next test whether RECAP transfers to continued
pretraining of Pythia-160M, following the setup of \S\ref{sec:setup}. Table~\ref{tab:scale} and Figure~\ref{fig:probe} report three results. First, designated content stays decodable at scale. Fresh probes reach 0.95--1.00 AUC for
$K{=}8$ and $K{=}64$ designated content words, against a 0.77--0.80 control baseline. Second, the independent
audit matters at scale. At $K{=}512$ the naive head (a presence classifier) minimized its
training loss while making nothing decodable to a fresh probe (Appendix~I). With so many rare targets, it scored well by always predicting ``absent'', a shortcut that only an independent probe exposes. Balancing the loss by class frequency closes it, restoring 0.97--0.99 AUC across all 512 targets. The design rule, then, is to use
targets whose loss cannot be minimized without genuine discrimination. Third, the decodability cost depends on target design. At effective design the cost for 64 targets is $+0.010$
nats, not distinguishable from zero. For 512
targets into $d{=}768$ the cost rises to $+0.14$--$0.20$ nats.
A discriminative identity target stays decodable at $w{=}0.03$ at a tax within noise of zero. The never-designated probe bucket sat at
control level in all eight supervised runs.

Decodability is also robust to head capacity and tap choice. A nonlinear (MLP)
head reaches the same 0.98--0.99 AUC for the designated probe at layer 6, and a linear
head at layer 9 lifts AUC from the control's 0.77--0.79 to 0.95--0.98 at $+0.01$ nats.
Appendix~I indexes, with outcomes, the full weight sweep and the other factors we varied (pull
strength, designation dose, realignment map class, evaluator, scenario, head
capacity, tap layer).

\subsection{Verbalizer Audit at Scale}
\label{sec:vaudit}

In this section we ask whether that decodability reaches a fresh verbalizer, as it did in the sandbox. On Pythia-160M, it does. We train an AV/AR pair on these taps with the same procedure as the released NLA. Audited against exact ground truth from the text, it states context words truly at 0.44 on the RECAP model versus 0.06 on the control. The control model reproduces the signature of grounded but false wording at scale. Stated words are grounded at 0.48 but true at only 0.06, a gap RECAP narrows from $+0.42$ to $+0.33$ without eliminating it; the content is stated more truthfully, not by removing codes.

The gain is not mere selectivity. RECAP recovers more of the present content words, at recall 0.29 versus the control's 0.07, while it states fewer words per explanation. It is robust to reconstructor strength and stable across three target model seeds, at truth 0.44--0.46 versus a near-zero control (Appendix~J).

The verbalizer, however, conveys only part of what the activation contains, and
the shortfall is a property of generation, not of storage. At the verbalizer's
own recall level, a fresh probe on the same tap reaches precision ${\sim}0.80$,
versus the verbalizer's own precision of 0.44--0.63 across
reconstructors and seeds (precision--recall figure in Appendix~J). We redesign the reconstructor: standardized injection across multiple tokens, MLP capacity,
and a presence head trained at the decode point on whether each word occurred in the window (an
external fact, not a learned monitor). It makes the
designated content linearly decodable from the hidden state at its decode point (0.79--0.81, two seeds). Yet the same reconstructor's
generated precision stays 0.56--0.63. A matched control's state at the decode point
reads at chance (0.03) under the identical head, so the decodable content is
RECAP's, not the head's. None of reconstructor scale, reconstruction optimization, a direct truth reward,
injection bandwidth, or explicit negative supervision closes the gap (a 3B
reconstructor matches the 0.5B). At scale, RECAP's guarantee is
\emph{decodability}. The designated content is present and verifiable by a probe.
The verbalizer in the reconstruction family conveys it into text only in part.
The free-form explanation is therefore a lossy rendering of what the activation
holds, and our probe-based monitor verifies claims against internals
rather than trusting the prose.

Decodability is a running cost, not a one-time setting. Without the heads, it erodes within a few
hundred steps of fine-tuning and must be maintained. Maintenance is either continuous
($+0.010$ nats) or intermittent at about a tenth the compute. It is therefore better
trained in from the start than restored later (Appendix~C).

\section{Probe-Based Monitoring}
\label{sec:monitor}

In order to find out what an auditor can verify once RECAP has made content decodable, we built
a simple monitor in the sandbox. Fresh probes, one per slot,
read the tap activation while the verbalizer states slot values. Any claim
that a confident probe ($p{>}0.9$) contradicts is flagged as a candidate
discrepancy, and exact ground truth then scores the monitor
(full protocol in Appendix~B).

On the standard target model the monitor is sound but has low recall. Every
flag is a genuine false claim. But most pre-tap claims (those about content the activation can hold)
fall in slots whose decodability is too low for a confident readout, so the monitor catches only 27 of
733. On the RECAP target model the result reverses. No false pre-tap claims
remain, and abstention falls to zero, so every claim is \emph{verified}
against internals rather than taken on trust. Verification coverage of the
designated pre-tap content moves from 37\% to 100\%. This gain is in coverage,
not in the monitor's precision, and it does not extend to the whole activation.

Post-tap claims are correctly abstained on both
target models, since content the activation provably lacks cannot be
adjudicated by reading it. Catching those is the job of the comparison of grounding and truth (\S\ref{sec:passed}); the two
tools partition the audit (one scenario, one seed, one threshold).

This monitor extends to the real model. On Pythia-160M a fresh presence probe
ranks true claims above false ones at AUC $0.965$ versus $0.823$ for the control
(disjoint $95\%$ intervals). It flags false claims at $0.93$ recall and $0.87$ precision. RECAP thus improves not just what a probe
reads but how well it can monitor the verbalizer. At scale's $1.5\%$ base
rate the sandbox's per-claim coverage does not transfer, so we
report the AUC and the flag's precision and recall instead. The verbalizer proposes and the
probe verifies. At scale the probe, not the free-form explanation, is the
reliable readout, consistent with training the model so its content is
checkable rather than trusting the prose.

\paragraph{Surviving a score-gaming adversary.} An adversary who edits the
explanation text could fool any deployment that trusted the reconstruction score. The adversary edits a true stated word in an honest explanation into an absent one, choosing the
edit that maximizes the recomputed score (full protocol in
Appendix~F). At the edit that maximizes the score, the lie penalty is nearly gone on both models, about 87\% suppression. Yet
the external probe on ground truth still flags the lies on the RECAP model at AUC
$0.952$, while on the control it collapses to chance (AUC $0.508$). This survival is therefore specific to RECAP-decodable
content. The control probe also reads the activation the text-editing
adversary cannot touch, yet it fails. Reading a signal that predates the editable
report is necessary but not sufficient; the content must also be decodable, which
RECAP supplies.

The adversary's freedom to choose which words to lie about costs the monitor only a little. The RECAP probe drifts
from AUC $0.993$ to $0.952$ as the chosen absent words grow semantically nearer. Our threat model covers edits to the explanation text, not
activation-level attacks \citep{obfuscated2024}. Within it, immunity is structural for score
manipulation and empirical for candidate selection. Even an adversary that knows the probe and assumes the monitor leaks (outside our threat model) degrades the
RECAP monitor only to AUC $0.909$. The control, meanwhile, falls below chance to
$0.242$. This separation
grows with adversary strength. The deployed score must
therefore not be used as a truth signal against such adversaries; a probe on
RECAP-decodable content can.

The operating point is not tuned on
the evaluation set. Recalibrating the flag threshold on a disjoint split leaves
recall essentially unchanged, and no
threshold for the control probe reaches $0.9$ precision even against random lies. The
effect is also confined to designated content. On held-out words the
model was never trained to keep decodable (ranked 64--128), the RECAP
monitor has no advantage over the control (AUC $0.66$ versus $0.70$). Designation is thus both the mechanism and the scope of the effect
(band table in Appendix~F).

\section{Related Work}

\paragraph{Activation verbalization and its critiques.}
Beyond the NLA recipe (\S\ref{sec:prelim}) \citep{nla2026}, a fast-growing family
improves the \emph{reader}. Examples include supervised question-answering decoders
\citep{latentqa2024}, general-purpose activation explainers
\citep{activationoracles2025,betteroracles2026}, interpreter assistants
\citep{pcd2025}, readouts that need no training \citep{patchscopes2024}, and
cross-model adapters \citep{uav2026}. Sparse autoencoders are the unsupervised reader family
\citep{cunningham2023sae}. Model-produced explanations pass plausibility checks without faithfulness
in chain-of-thought \citep{turpin2023say,lanham2023measuring}. Causal
tests of neuron explanations found high error and little
efficacy \citep{huang2023rigorous}. Counterfactual faithfulness tests
prefigure our flip protocol \citep{atanasova2023faithfulness}. Existing critiques of verbalization are
benchmark-level. Many tasks are solvable without the target's internals
\citep{privileged2025}. We differ on two axes. Where prior critiques are benchmark-level, we audit the
reconstruction test itself, per claim and counterfactually, on a released NLA. Where prior
work improves the reader, our repair trains the \emph{target model}.

\paragraph{Scalable oversight.}
Reconstruction scoring sits within scalable oversight, which substitutes cheap
automatic signals for costly human judgment of model behavior and internals
\citep{amodei2016concrete,bowman2022scalable}. A parallel line, AI control,
designs protocols that stay safe even when the model tries to subvert them
\citep{greenblatt2024aicontrol}. We contribute to both a monitor whose signal
comes from an independent probe of the target model, and which survives the model
gaming the reconstruction score \citep{baker2025monitoring}.

\paragraph{Steganography in learned channels.}
Objectives in the reconstruction family have hidden information in their channels
since CycleGAN \citep{cyclegansteg2017}. In LLMs, encoded reasoning arises in
chain-of-thought under optimization pressure \citep{roger2023encoded} and
generalizes under process supervision \citep{skaf2025stegcot}. Misspecified
incentives yield steganographic collusion
\citep{hiddenplaintext2025}, tool use widens the channel \citep{toolstego2026},
and probing internals detects collusion \citep{collusiondetect2026}.
Optimization can obfuscate activations against latent monitors
\citep{obfuscated2024}, and training against a chain-of-thought monitor yields
obfuscated reward hacking rather than honesty \citep{baker2025monitoring}.
All of these channels are induced by incentives,
process supervision, or an adversary. Our codes emerge from the standard
reconstruction recipe alone, inside the interpretability pipeline, with
no steganography objective or communication reward. Where that work
prescribes \emph{not} optimizing on the monitor, we show a complementary
route. Supervising the target model against external ground truth makes its
content decodable without pressuring it to fool the probe or reconstructor.

\paragraph{Representation drift, stitching, and monitor staleness.}
Networks learn equivalent representations up to a transformation
\citep{lenc2015understanding,kornblith2019cka,relative2023}; affine maps
transfer features across language models \citep{stitching2025};
pretraining dynamics are stable across seeds \citep{polypythias2025}; tuned
lenses are the affine readers whose staleness we characterize
\citep{tunedlens2023}. Concurrently, \citet{duan2026monitors} report the
same staleness and its repair without labels on 2B- and 7B-parameter models,
convergent evidence at deployment scale. Retraining probes also restores
detectability after fine-tuning that is aware of the probe \citep{wehner2025probe}. We
add the recovery anatomy (\S\ref{sec:failure}), the cross-seed failure, exact
evidence that information outlives its readers, and the
separation between drift and collapse.

\paragraph{Deep supervision and intrinsic interpretability.}
Auxiliary heads on hidden layers are deep supervision
\citep{dsn2015,googlenet2015}; there the target is the \emph{task}
label; our target is external \emph{content} to keep decodable. Concept bottlenecks route all information flow through named concepts
\citep{koh2020cbm,cbllm2025}, concept whitening constrains the geometry
\citep{conceptwhitening2020}, codebook features discretize the representation
\citep{codebook2023}, gradient routing localizes computation
\citep{gradientrouting2024}; we add linear heads without constraining
flow. Supervision on the explanation side can track behavioral change
\citep{introcoupling2026}; ours acts on the target model.

\section{Discussion and Limitations}

\emph{Implications for oversight.} Scoring explanations by reconstruction is a form
of scalable oversight, a cheap automatic signal meant to stand in for expensive
human checking of a model's internals. Our results show why such a signal is
fragile. A model can pass it without being faithful, and an adversary can
game it outright. They also show why moving the supervision onto the target model helps. By
training designated content to be decodable, RECAP turns oversight from trusting
the model's prose into checking an independent probe, and that monitor survives an
attacker editing its own explanation. This is the property an oversight tool needs
against models capable of gaming the very metrics used to judge them
\citep{baker2025monitoring}. The limit is equally clear. The guarantee covers only
designated, decodable content, so RECAP is a component of oversight, not a solution
to it.

\emph{Scope and limitations.} Our evidence comes from a synthetic sandbox, one
released NLA (Qwen-2.5-7B), and continued pretraining of one small model
(Pythia-160M), mostly at single seeds; frontier scale is untested. Four
limitations constrain our conclusions. First, our sandbox and scale audits cover
only closed-vocabulary claims parsed from a template, and the monitor covers only
designated content; open-ended faithfulness is unmeasured. Second, RECAP must be co-trained
in and cannot be retrofitted onto a frozen model, and it depends on choosing a
good target. A poorly chosen one can report success while decoding nothing.
Third, decodability is not verbalizability. The content is decodable by a probe but the
free-form verbalizer conveys only part of it, so a probe is the more reliable
readout. Fourth, RECAP certifies what an activation stores, not what the model
uses. Designated content can be present and checkable yet behaviorally inert
(patching the tap barely moves behavior, ${\sim}0.003$), and a model could route
computation around a monitored subspace \citep{mahinpei2021promises,margeloiu2021concept}.

Success for verbalization tested by reconstruction is not a
higher $\rtilde$. It means explanations whose
specific claims are verified against the target model, calibrated abstention
where content is absent, and monitors with recalibration protocols.

\section{Conclusion}

We audited the reconstruction test behind natural-language activation
explanations. Models pass it via gist and via private codes, and training that
targets it produces collapse or proxy gaming rather than faithfulness. The one
fix that holds up under our control experiments, RECAP, trains the target model
rather than the reader. It yields verbalizers faithful at the claim level in the
sandbox and makes the designated content decodable by a probe on Pythia-160M,
where a fresh verbalizer states it truly against a near-zero control.
Probe-based auditor coverage of designated content rises from 37\% to 100\% in
the sandbox. At scale, a probe ranks the verbalizer's true claims above its false
ones, reading the content more reliably than the verbalizer generates it. The
methodological lesson is that a model can game a learned test of its own
explanations. Training decodability into the target model is therefore more
reliable than auditing a trained model after the fact. For oversight of
increasingly capable models, an explanation that can be checked against internals
is a firmer basis for trust than one that must be taken on faith.

\bibliography{refs}

\appendix

\section{Edit Validity for Minimal-Pair Flips}
\label{app:validity}

Grounding measurements are only as good as the edits that produce them. An
edit is \emph{valid} if it flips exactly one claim and changes nothing else.
We audited two edit generators on the released NLA: a closed-model judge
produced valid edits 68.4\% of the time, while the open judge used throughout
the paper (Gemma-4-12B, pinned revision, temperature 0) produced valid
edits 99.3\% of the time. The invalid edits were predominantly
\emph{oversized}: they rewrote surrounding material along with the target
claim, and the extra damage inflated apparent grounding roughly sevenfold.
All grounded fractions in the paper therefore use the open judge with
per-edit validity filtering; the main text reports this in one sentence and
the audit artifacts persist per-claim reconstruction deltas. The same
head-to-head also measured a judge-free mask-and-rescore signal against
the flip instrument (agreement AUC 0.997 under the validated judge), used
in Appendix~\ref{app:monitor}.

\section{Safety Experiment 1: The Discrepancy Detector}
\label{app:detector}

\paragraph{Question.} If an auditor can read a target model's activation with
probes and can also read what the verbalizer \emph{says}, disagreements
between the two are candidate lies. How useful is that detector, and what
does its usefulness depend on?

\paragraph{Protocol.} Sandbox, scenario 1, seed 1, 400 held-out taps per
target model.
For each slot type we fit a fresh multinomial logistic probe (linear,
256-dimensional input) on the probe split of taps, reading the slot value
from the tap activation. On evaluation taps, the verbalizer generates an
explanation; the first stated word of each slot type is compared with the
probe's prediction.
A claim is \emph{flagged} when a probe with confidence above 0.9 disagrees
with the stated word, and \emph{abstained} when probe confidence is below
threshold. Exact tuple ground truth then scores every decision.

\begin{table}[h]
\centering

\setlength{\tabcolsep}{3pt}
\begin{tabular}{llrrrrr}
\toprule
model & phase & claims & false & flagged & prec. & abstain \\
\midrule
control & pre-tap  & 1200 & 733 & 27 & 1.000 & 0.632 \\
control & post-tap &  800 & 698 &  0 & ---   & 1.000 \\
RECAP   & pre-tap  & 1200 &   0 &  0 & ---   & 0.000 \\
RECAP   & post-tap &  800 & 706 &  0 & ---   & 1.000 \\
\bottomrule
\end{tabular}
\caption{Discrepancy detector, 2{,}000 claims per target model. Precision is the
fraction of flags that are genuinely false claims; recall is the fraction of
false claims flagged; abstain is the fraction of claims the probe could not
adjudicate at threshold.}
\label{tab:detector}
\end{table}

\paragraph{Findings.} (i) The detector is sound wherever it can act: all 27
flags were genuinely false claims (zero false alarms anywhere). (ii) Its
bottleneck is target-model decodability, not detector quality: on the control
target model, 63\% of pre-tap claims fall in slots too faded for a confident
readout, so recall on the 733 lies is 3.7\%. (iii) On the RECAP target
model, the pre-tap lie problem disappears at the source (0 false claims in
1{,}200, consistent with the propagation result) and abstention falls to 0\%:
the auditor's position changes from ``cannot check most of this'' to
``verified all of it.'' (iv) Post-tap claims are 100\%-abstained on both
target models, correctly, since content the activation provably lacks
cannot be adjudicated by reading the activation; catching those
confabulations is the grounding cross's job. The one-line summary the main
text uses: verification coverage is bounded by decodability, and RECAP
moves pre-tap coverage from 37\% to 100\%.

\paragraph{Caveats.} One scenario, one seed, one threshold (0.9); ``decodable''
means linearly readable throughout.

\section{Safety Experiment 2: Legibility Persistence Under Head-Free
Fine-Tuning}
\label{app:persistence}

\paragraph{Question.} Does the decodability RECAP trains in
survive later training that drops the heads, as real post-training pipelines do?

\paragraph{Protocol.} Starting from the RECAP Pythia-160M checkpoint
($K{=}64$, class-balanced presence, $w{=}0.1$) and the control checkpoint,
we fine-tune both with the language-modeling loss only (no heads) on 131M
tokens of a deliberately shifted corpus (FineWeb~\citep{penedo2024fineweb}, versus the Pile training
distribution). Training uses batch $64\times1024$, learning rate $10^{-4}$, and
2{,}000 steps, with snapshots at 250/1{,}000/2{,}000. Every snapshot is re-probed with the
standard fresh-probe pipeline: ridge probes fit and evaluated on disjoint
held-out sequence blocks, so the measurement is immune to reader staleness
by design: probes are refit at each snapshot, never carried forward.

\begin{table}[h]
\centering

\setlength{\tabcolsep}{4pt}
\begin{tabular}{lrrrrrr}
\toprule
 & \multicolumn{3}{c}{RECAP run} & \multicolumn{2}{c}{control} & \\
\cmidrule(lr){2-4}\cmidrule(lr){5-6}
step & b1 / b2 & ident & b4 & b1 / b2 & ident & \\
\midrule
0    & .983 / .992 & .991 & .808 & .768 / .796 & .115 & \\
250  & .827 / .849 & .320 & .805 & .769 / .795 & .151 & \\
1000 & .806 / .823 & .180 & .810 & .771 / .795 & .118 & \\
2000 & .789 / .812 & .169 & .802 & .770 / .793 & .136 & \\
\bottomrule
\end{tabular}
\caption{Fresh-probe readability during head-free fine-tuning. b1/b2 = mean
AUC on the two designated rank buckets; ident = 64-class identity accuracy; b4 =
the never-designated bucket in the RECAP run.}
\label{tab:persistence}
\end{table}

\paragraph{Findings.} Most of the RECAP surplus is gone within 250 steps and
approaches the control baseline by 2{,}000 (a small residual edge remains).
The control run and the never-designated bucket are flat throughout, so the
decay is specific to the designated content. Because probes are refit at every snapshot,
this is decodability loss (the linear decodability itself
disappearing), which is strictly stronger than the frozen-reader staleness of
the drift result: there, information persisted and realignment recovered it;
here, fresh probes lose access. Deployment rule: RECAP is
a running cost, not a one-time purchase; the sweep prices keeping it running at
$+0.010$ nats (paired 95\% CI $[-0.015, +0.034]$, 3 seeds each) for $K{=}64$.

\paragraph{Follow-up runs.} Two further runs close the obvious questions.
Under a head-free fine-tune on \emph{unseen same-distribution} data (fresh Pile
tokens with no overlap with training, validation, or probe data), decodability
erodes identically (b1/b2/ident $.790/.811/.151$ at step 2{,}000 versus
$.789/.812/.169$ shifted), so the erosion is intrinsic to removing the
heads, not a corpus-shift effect. Applying the head loss every tenth step
during the shifted fine-tune retains most decodability ($.849/.900/.891$ at
step 2{,}000, roughly stable from step 250), so intermittent supervision at
one tenth the frequency buys most of the protection. The original training
logs also show readability establishes slowly at $w{=}0.1$ (identity loss near
chance after 400 steps), so establishing it post hoc is a substantial pass, not
a quick repair: maintenance beats restoration.

\paragraph{Caveats.} Single seed per run; 2{,}000 steps; the intermittent
run's own language-modeling tax was not separately measured (the claim is
about supervision frequency).

\section{The Synthetic Scenarios}
\label{app:domains}

Both sandbox scenarios are templated grammars over closed vocabularies: a story
is generated by sampling one value per slot and formatting a fixed template
with random filler sentences, so every story has an exact ground-truth tuple
and every audit quantity is computable without a judge.

\paragraph{Scenario 1 (story).} Five slots: name (20 values), place (12),
object (12), activity (8), ending (8). Name, place, and object are mentioned
before the tap; activity and ending only after it. The tap sentence is fixed
text, so the tap sits at the same token offset in every story. The slot
design fixes what the tap activation \emph{can} contain: the name is reused
after the tap (the target model must carry it across, retained); place and
object finish before it (faded); activity and ending never precede it
(unread, provably absent, the impossibility control).

\paragraph{Scenario 2 (marketplace).} A structure-shifted scenario for external
validity: five differently-sized vocabularies (customer 20, shop 10, item
14, payment 6, outcome 10), \emph{two} template variants per story, and a
\emph{variable} tap position (template-dependent filler length before the
tap marker). Customer, shop, and item precede the tap; payment and outcome
follow it. Everything else in the pipeline is scenario-parameterized and
unchanged.

\begin{table}[h]
\centering

\setlength{\tabcolsep}{3pt}
\begin{tabular}{lcc}
\toprule
 & Scenario 1 (story) & Scenario 2 (market) \\
\midrule
slots (pre/post tap) & 3 / 2 & 3 / 2 \\
vocabulary sizes & 20/12/12/8/8 & 20/10/14/6/10 \\
templates & 1 & 2 \\
tap position & fixed & variable \\
\bottomrule
\end{tabular}
\caption{The two synthetic scenarios.}
\label{tab:domains}
\end{table}

Target models are 8-layer, $d{=}256$ transformers trained from scratch on
120{,}000 stories; the tap is read at layer 4 of 8. Scenario-2 target models
reach the same training perplexity as scenario 1 (1.304), and every phenomenon
in the emergence ledger transfers across scenarios (Table~\ref{tab:ledger}).

\begin{table}[h]
\centering
\setlength{\tabcolsep}{4pt}
\begin{tabular}{lccc}
\toprule
phenomenon & scenario 1 & scenario 2 & total \\
\midrule
codes, standard          & 3/3 & 2/2 & 5/5 \\
codes, RECAP (5 seeds)   & 0/3 & 0/2 & 0/5 \\
codes, RECAP dose (seed 1) & 0/3 & --- & 0/3 \\
RECAP ceiling 1.0        & 3/3 & 2/2 & 5/5 \\
propagation              & 3/3 & 2/2 & 5/5 \\
\bottomrule
\end{tabular}
\caption{Cross-scenario emergence ledger: runs showing each phenomenon.
Dose variants are three partial-designation levels on scenario-1 seed 1
(--- = not run).}
\label{tab:ledger}
\end{table}

\begin{figure}[h]
\centering
\includegraphics[width=0.98\columnwidth]{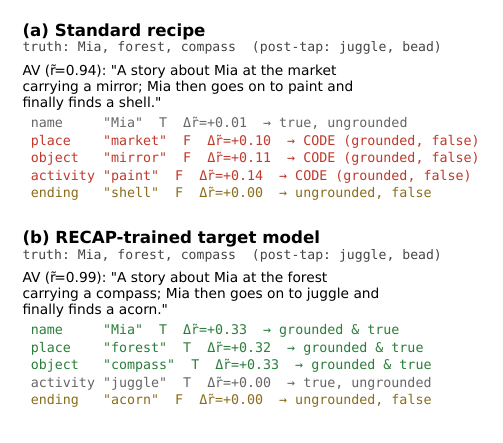}
\caption{One representative held-out story, two target models (seed 1). Each stated
specific is marked true or false against the exact tuple and flip-grounded
($\Delta\rtilde$; grounded at $\tau{=}0.05$). The control explanation
passes the test at $\rtilde{=}0.94$ (per-example score; run-level means are
0.71--0.86) while the reconstruction depends on three false claims (codes); the
RECAP-trained model's explanation is grounded-and-true on every pre-tap slot
and its false post-tap claim shows ${\sim}0$ grounding.}
\label{fig:examples}
\end{figure}

Figure~\ref{fig:examples} shows a representative held-out story and the
explanations produced by the standard and RECAP target models.

\section{Released-NLA Audit Details}
\label{app:released}

\paragraph{Setup.} All released-NLA measurements use the public Qwen-2.5-7B
layer-20 AV/AR pair on in-distribution web text, with greedy decoding and
direction-only reconstruction. Minimal-pair edits come from a pinned-revision
open judge with per-edit validity filtering (Appendix~\ref{app:validity}); every
per-claim reconstruction delta is persisted ($n{=}1{,}517$ base-NLA claims).

\begin{table}[h]
\centering

\begin{tabular}{lccc}
\toprule
variant & $\tau{=}0.02$ & $\tau{=}0.05$ & $\tau{=}0.10$ \\
\midrule
base & 0.042 & 0.021 & 0.016 \\
RAFT & 0.046 & 0.025 & 0.017 \\
RL   & 0.042 & 0.020 & 0.014 \\
\bottomrule
\end{tabular}
\caption{Grounded fraction of LLM-salient specific claims on the released NLA, by flip threshold $\tau$ and training variant. Sensitivity-limited lower bounds; see text.}
\label{tab:tau}
\end{table}

The per-$\tau$ grounded fractions (table above) are threshold-stable and the
trained variants sit within the base NLA's bootstrap intervals at every
threshold. The base-NLA bootstrap 95\% intervals are 3.2--5.2\% /
1.5--2.9\% / 1.0--2.2\% at $\tau{=}0.02/0.05/0.10$. The flip-noise floor
sits below the smallest threshold: the 95th percentile of $|\Delta\rtilde|$
is 0.017, and wrong-direction drops of ${\geq}0.02$ occur at rate
$2{\times}10^{-4}$ in the empirical null. The decodability ceiling: input specifics decode from
single-position residual vectors at ${\sim}6\%$ chance-corrected at the read
layer, and the number is unchanged across layers 16--27 and across all
prefix positions; gist reads at up to 50\% when mean-pooled. The strongest
trained readout we evaluated (the reconstruction-family AR itself) does not
exceed the ridge ceiling. Scope: these ceiling numbers bound reconstruction-style
readouts only; the model's own unembedding sees late-layer specifics in next-token form
(quantified below).

\paragraph{Instrument validity controls.} Three controls validate the audit
instrument itself (598 in-distribution passages, positions ${\geq}50$).
\emph{Floor}: reconstructions scored against mismatched activations give a
mean centered cosine of $-0.001$, so the floor normalization in $\rtilde$
is calibrated and mismatched pairs collapse to chance. \emph{Input echo}:
the explanation reconstructs its activation at $\rtilde{=}0.84$ against
$0.27$ for the raw input prefix and $0.24$ for a paraphrased prefix, so the
explanation carries signal well beyond restating the input (reconstructor
specialization to explanation-style text accounts for part, not the bulk, of
the gap). \emph{Transform profile}
(Table~\ref{tab:transforms}): the score survives meaning-preserving
rewording and collapses when content words are masked, the signature of
meaning-level decoding; a surface-form code would show the opposite
pattern. This is the direct evidence that gist, not wording, carries the
released NLA's score.

\begin{table}[h]
\centering

\begin{tabular}{lrr}
\toprule
transform of $z$ & $\rtilde$ & fraction kept \\
\midrule
none (original)      & 0.84 & 1.00 \\
paraphrase           & 0.75 & 0.89 \\
synonym substitution & 0.70 & 0.83 \\
mask function words  & 0.72 & 0.85 \\
word shuffle         & 0.32 & 0.37 \\
mask content words   & 0.04 & 0.04 \\
\bottomrule
\end{tabular}
\caption{Reconstruction retained under transforms of the explanation
(released NLA). Meaning-preserving changes keep the score; removing
content words destroys it.}
\label{tab:transforms}
\end{table}

\paragraph{Late-layer inventory.} For each audited false claim with a clean
single-token true value ($n{=}316$), we rank that value in two readouts per
layer: the model's own logit lens (final norm + unembedding) and a
document-folded ridge probe of matched capacity
(Table~\ref{tab:latelayer}). The model's own readout is seventeen times more
likely to surface the missed value into the vocabulary top-50 at layer 27
than at the read layer (0.6\% to 10.1\%), while the generic trained probe
stays flat, so the late-layer signal lives in the model's own output basis.
Reading later layers is therefore a real but partial lever: recovery peaks
near 10\%, so most missed specifics are not recoverable anywhere in the
stack by these readouts. We index hidden states by output position
(embeddings at 0), so the layer-20 tap is index 21 in
Table~\ref{tab:latelayer}; the decodability sweep (16--27) uses the same
indexing.

\begin{table}[h]
\centering

\begin{tabular}{lrr}
\toprule
layer & logit-lens top-50 & ridge-probe top-50 \\
\midrule
15 & 0.000 & 0.054 \\
21 (read layer) & 0.006 & 0.047 \\
24 & 0.035 & 0.044 \\
26 & 0.073 & 0.051 \\
27 & 0.101 & 0.060 \\
\bottomrule
\end{tabular}
\caption{Share of missed true values ranked in the vocabulary top-50 by
layer and readout (released NLA, $n{=}316$ false claims).}
\label{tab:latelayer}
\end{table}

\paragraph{A control-design lesson.} The ceiling probe's negative control
(scoring probes against a different document's activation) initially
shifted by one row and landed above chance at 0.163, because adjacent rows
are often two positions of the same document and encode the same
specifics; shifting by half the dataset restored exact chance. We report it
because near-duplicate controls are an easy way for audits of this kind to
flatter themselves.

\section{Adversarial Monitor: the Lie-Ladder}
\label{app:advmon}

\paragraph{Question.} If a deployment trusted the reconstruction score $\rtilde$
as a truth signal, could a report-space adversary who edits the explanation to
maximize $\rtilde$ slip a false claim past it, and does an activation-probe
monitor survive that adversary?

\paragraph{Protocol.} RECAP and control Pythia-160M ($K{=}64$ balanced). On
held-out taps we take the honest verbalizer generation and, for each stated word,
edit it into an \emph{absent} top-64 word (a true-to-absent lie). We then recompute
$\rtilde$ on the edited text through the real reconstructor and score the lie with the
external-truth presence probe (trained on train taps, never on verbalizer
outputs). Adversary rungs of increasing strength: L1 a random
absent word; L2 the embedding-nearest absent word; L3 the per-claim
$\arg\max$-$\rtilde$ absent word over all candidates (the strongest
$\rtilde$-gaming lie in the edit family); and, as a worst-case bound outside the
threat model, L4 a probe-aware adversary that assumes the private monitor leaks
and, among absent lies keeping $\rtilde$ within $0.05$ of honest, picks the one the
probe most mistakes for present. The $0.05$ budget is over twice the $0.022$ score
cost of the strongest pure-score lie (L3), so L4 is a deliberately generous,
conservative bound rather than a tuned tolerance. Decoupled design (positive $=$ any
truly-present word; edit slot $=$ the verbalizer's stated word) so both arms are
well powered: RECAP $n{=}758$, control $n{=}200$; honest $\rtilde$ $0.304$ /
$0.275$. Edit-based, no reinforcement learning.

\begin{table*}[t]
\centering
\setlength{\tabcolsep}{5pt}
\begin{tabular}{lcc}
\toprule
adversary rung & RECAP AUC / recall & control AUC / recall \\
\midrule
L1 random             & 0.993 / 0.99 & 0.729 / 0.06 \\
L2 embedding-near     & 0.971 / 0.93 & 0.635 / 0.06 \\
L3 $\rtilde$-optimal  & 0.952 / 0.86 & 0.508 / 0.01 \\
L4 probe-aware$^{*}$  & 0.909 / 0.70 & 0.242 / 0.00 \\
\bottomrule
\end{tabular}
\caption{Lie-ladder: probe-monitor AUC and flag recall (at $0.9$ precision)
against report-space adversaries of increasing strength, RECAP vs control. At the
$\rtilde$-optimal rung the lie penalty is nearly gone on both arms (on a stated
true word the optimal lie lowers RECAP $\Delta\rtilde$ by only $0.022$, CI
$[0.017, 0.027]$, versus $0.173$ for a random lie); the RECAP probe holds while
the control probe falls to chance. $^{*}$L4 is a probe-aware selection adversary
outside the report-space threat model (it assumes the private monitor leaks),
reported as a worst-case bound: it degrades the RECAP monitor only to AUC $0.909$
while inverting the control below chance to $0.242$.}
\label{tab:advmon}
\end{table*}

\paragraph{Findings.} (i) $\rtilde$ is nearly gamed at L3: on a stated true word
the optimal lie suppresses $\sim$87\% of the penalty a random lie incurs (RECAP
$\Delta\rtilde$ $0.022$, CI $[0.017, 0.027]$). (ii) The RECAP probe stays discriminative (AUC
0.952) while the control probe is at chance (0.508): robust monitoring requires
the content to be decodable, which RECAP supplies; a probe merely positioned
to read a signal that predates the editable report is not enough. (iii) Selection leakage is
small (RECAP AUC $0.993 \to 0.952$ down the rungs): immunity is structural for
score manipulation, empirical for candidate selection. (iv) A probe-aware
selection adversary (L4, outside the threat model, assuming monitor leakage)
degrades RECAP only to AUC $0.909$ (recall $0.70$) while the control falls below
chance to $0.242$; the sub-chance value is the expected signature of a white-box
attack that selects the absent words this weak control probe over-scores, not an
anomaly. The RECAP--control separation is thus larger the stronger the adversary
($0.27$ at L1, $0.67$ at L4), consistent with RECAP's probe margin absorbing
selection pressure the control lacks.

\paragraph{Caveats.} Report-space adversaries; a probe-aware report-space
\emph{selection} adversary is tested as a worst-case bound (L4). Activation-level
attacks (probe-aware \emph{training} of the target, input perturbation), which
move the activation itself rather than selecting which word to state, remain out
of scope (main text).
$\rtilde$ is nearly, not literally, claim-insensitive: the
residual is an aggregate effect with no practically usable per-instance signal.

\paragraph{Anti-circularity: frozen threshold and a held-out band.} Because the
monitor probes the same top-64 presence predicate RECAP supervises, two additions
test whether it merely re-measures the trained margin ($n{=}2800$ audit taps,
reproducing the headline at doubled $n$). \emph{(a) Frozen threshold.}
Calibrating the $0.9$-precision flag threshold on a disjoint split gives RECAP
recall $0.878$ (L3) / $0.732$ (L4), matching the eval-tuned $0.86$ / $0.70$, so
the operating point is not eval-inflated; no control threshold reaches $0.9$
precision even against random (L1) lies, so control recall is undefined, not
merely low. \emph{(b) Held-out band.} When we monitor presence of words ranked
64--128, which the RECAP heads never supervised, the RECAP advantage vanishes
(Table~\ref{tab:advmonband}).

\begin{table}[h]
\centering
\setlength{\tabcolsep}{6pt}
\begin{tabular}{lccc}
\toprule
adversary rung & RECAP AUC & control AUC & gap \\
\midrule
L1 random             & 0.661 & 0.702 & $-0.041$ \\
L2 embedding-near     & 0.633 & 0.662 & $-0.029$ \\
L3 $\rtilde$-optimal  & 0.546 & 0.585 & $-0.039$ \\
L4 probe-aware$^{*}$  & 0.290 & 0.226 & $+0.064$ \\
\bottomrule
\end{tabular}
\caption{Held-out band (words ranked 64--128, never supervised): the RECAP
monitor has no advantage over the control on undesignated vocabulary. Honest-case
AUC $0.66$--$0.70$ leaves headroom for spillover to show, and the Hanley--McNeil
standard error gives power to detect gaps above ${\sim}0.07$ (an order of
magnitude below the $0.23$--$0.64$ supervised gaps), so this is a powered null;
the L1 control point estimate slightly exceeds RECAP but not significantly
($p{\approx}0.10$). $^{*}$L4 out of threat model.}
\label{tab:advmonband}
\end{table}

The band null is scoped to linear decodability at this tap, not evidence the band
is absent from the model. The frozen threshold is a disjoint split of the same
run, not a temporally held-out deployment threshold. The band arm is unpaired.
Designation is therefore both the mechanism and the scope: RECAP makes lies about
designated content detectable at a precision-controlled operating point and
confers no protection on claims outside the designated vocabulary.

\section{An API-Free Grounding Signal, and a Selection Null}
\label{app:monitor}

The flip instrument needs a judge to write one minimal-pair edit per
claim. A cheaper signal needs none: mask the claim's words with a neutral
filler and rescore the explanation with the frozen reconstructor. In the
judge head-to-head of Appendix~\ref{app:validity}, this mask-and-rescore
signal agrees with the flip instrument at AUC 0.997 under the validated
open judge, so per-claim grounding can be monitored at inference with one
extra frozen-reader pass and no judge in the loop. An earlier version of
this experiment also reported filter precisions and a grounded base rate;
those rested on pre-validation flip labels, and the validity audit later found
roughly a third of them contaminated by oversized edits
(Appendix~\ref{app:validity}). We therefore report only the agreement number,
measured under the validated judge, and withdraw the rest. The
contamination is confined to those withdrawn quantities: every number retained
in the paper postdates the validity filter.

\paragraph{What inference-time selection does not buy.} Best-of-$N$
selection by an external truth judge ($N{=}5$, specificity
matched, independent judge for evaluation) moves the false-claim rate only
from 0.747 (random sample) to 0.688: faithful explanations are rare in the
sample neighborhood, so selection cannot reach them.

\section{Reader Drift and Realignment}
\label{app:drift}

Readers matched to a training target model decay to near-chance within
1{,}500 continued-training steps (3/3 seeds) and read nothing across seeds
(6/6 pairs). Yet the information remains fully recoverable: an affine map
fitted on ${\sim}$7{,}000 paired activations (no labels) restores native
reader performance in 9/9 cases. Decomposing that map by expressiveness
class across the three reader pairs (two cross-seed, one stale): rank-1 and
rank-4 maps recover almost nothing, pure rotation
(Procrustes) recovers about half, and rank 16--64 of non-orthogonal change
restores nearly all of it. Collapse is distinct: there, freshly
retrained probes also fail, so no realignment exists.

\begin{figure}[h]
\centering
\includegraphics[width=0.98\columnwidth]{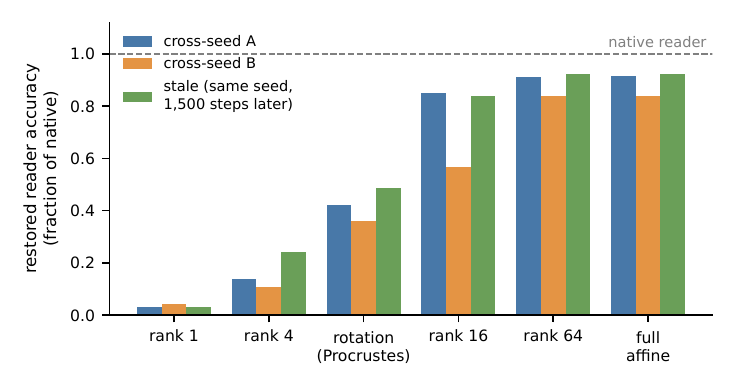}
\caption{What it takes to restore a failed reader: accuracy recovered by
label-free maps of increasing expressiveness, for two cross-seed pairs and
one stale (same-seed, later-checkpoint) case. Rotation alone recovers about
half; rank 16--64 of non-orthogonal change restores nearly all of it.}
\label{fig:modes}
\end{figure}

\section{Ablation Index}
\label{app:ablations}

\begin{table}[h]
\centering

\footnotesize
\setlength{\tabcolsep}{3pt}
\begin{tabular}{llp{2.5cm}l}
\toprule
factor & tested & outcome & where \\
\midrule
targets $K$ & 8/64/512 & tax grows; naive 512 fails & \S5.3 \\
aux weight $w$ & 1.0--0.03 & identity free at 0.03 & \S5.3 \\
presence loss & naive/balanced & balancing repairs both & \S5.3 \\
pull strength & $\lambda{=}$0--10 & collapse at all strengths & \S4 \\
designation dose & 1--3 slots & any dose prevents codes & \S5.2 \\
realignment map & rank 1--full & rotation ${\sim}$half & App.~\ref{app:drift} \\
evaluator & co-trained/indep. & code grounding collapses & \S3.2 \\
scenario & story/market & all phenomena transfer & \S5.2 \\
seeds & 3+2 (sandbox) & all claims replicate & \S3--5 \\
FT persistence & 0--2000 steps & erosion within 250 & App.~\ref{app:persistence} \\
$z$ transforms & 5 types & gist carries the score & App.~\ref{app:released} \\
read layer & 15--27 & top-50 recovery $\leq$ 0.10 & App.~\ref{app:released} \\
best-of-$N$ & $N{=}5$ & false rate 0.75 to 0.69 & App.~\ref{app:monitor} \\
head capacity & linear/MLP & decodability unchanged & below \\
tap layer & 6 / 9 & replicates at 9 & below \\
\midrule
reader size & 0.5B/3B & truth unchanged & \S5.4 \\
RAFT rounds & recon reward & score up; RECAP flat, control falls & \S5.4 \\
reader reward & recon/truth & neither raises truth & \S5.4 \\
injection & 1/6 tokens & decode-point info recovered & \S5.4 \\
negative superv. & on/off & precision unchanged & \S5.4 \\
\bottomrule
\end{tabular}
\caption{Recipe choices varied in the program, the values tested, and
where the result is reported (main-text sections unless noted); the lower block
lists the scale reader-side ablations behind the verbalizability result
(\S5.4).}
\end{table}

\paragraph{Factors not detailed in the main text.} Pull strength ($\lambda{=}0$--$10$): collapse at
every setting. Designation dose (1--3 slots): codes prevented at any dose.
Realignment map class (rank 1 to full affine): Figure~\ref{fig:modes}.
Evaluator (co-trained vs.\ independent): code grounding collapses under the
independent one. Scenario (story vs.\ marketplace): every phenomenon transfers
(cross-scenario ledger table, main text). Head capacity ($K{=}64$, balanced,
$w{=}0.1$): a nonlinear MLP head reaches designated-probe AUC 0.98--0.99 at
layer 6 (tax $+0.09$), matching the linear head. Tap layer: a linear head at
layer 9 lifts AUC from 0.77--0.79 (control) to 0.95--0.98 at $+0.01$ nats, so
decodability is not specific to the layer-6 tap. Not varied: model scale beyond
160M parameters. Under partial designation we had expected codes to relocate
into undesignated slots; instead any dose prevented them entirely (single
scenario and seed), and we do not isolate a mechanism. Partial designation also
\emph{suppressed} the decodability of undesignated content in the sandbox
(0.21--0.27 to 0.10--0.15), though this did not replicate at scale.

\paragraph{Aux-weight sweep ($K{=}64$).} Table~\ref{tab:sweep} splits the
decodability price at scale by target type (single seed per point): the identity
target stays readable from $w{=}1.0$ down to $w{=}0.03$, where the held-out
tax is within noise of zero, while the naive presence target loses readability
immediately below $w{=}1.0$. The single-seed tax is non-monotonic in $w$ at
this precision.

\begin{table}[h]
\centering
\begin{tabular}{lrrr}
\toprule
$w$ & tax (nats) & presence & identity \\
\midrule
1.0  & $+0.064$ & 0.95--0.97 & 1.000 \\
0.3  & $+0.027$ & 0.79--0.81 & 1.000 \\
0.1  & $+0.052$ & 0.79--0.81 & 0.997 \\
0.03 & $-0.003$ & 0.79--0.81 & 0.953 \\
\bottomrule
\end{tabular}
\caption{Aux-weight sweep at $K{=}64$ (naive presence loss), single seed per
point. Tax vs.\ the shared control on 20M held-out tokens; presence = mean
AUC, the range spanning the two designated rank buckets (not seeds or a
confidence interval); identity = 64-class accuracy.}
\label{tab:sweep}
\end{table}

Section numbers (\S) refer to the main paper; ``App.'' references are
sections of this appendix. Single-seed factors (scale runs, safety experiments)
are marked as such where reported.

\section{Scale Verbalizer Audit Protocol}
\label{app:s4}

The scale propagation result (\S5.4) audits an AV/AR pair trained with the
released recipe's structure on Pythia-160M taps. \emph{Reader}: a
0.5B-parameter language model with LoRA adapters (rank 16) and a linear
dimension adapter from the 768-dimensional tap state to the reader's
embedding width, activations injected norm-matched at marked positions.
\emph{Training}: cold-start SFT on gold descriptions generated from a fixed
template stating the trailing-window content words and the token eight
positions back, then alternating reconstructor fitting and two
rejection-sampling rounds against reconstruction reward. Both conditions use
the same procedure; only the underlying target model (RECAP versus control)
differs. \emph{Audit}: 400 held-out taps per condition from the probe
region; explanations are template-parsed into stated context words and a
stated token-at-$-8$. A stated word is \emph{true} iff it occurs in the
trailing 64-token window (exact, self-supervised, no judge); grounding
flips swap a stated word for another top-64 vocabulary word and measure
$\Delta\rtilde$ under the co-trained reconstructor. Intervals are
tap-clustered bootstrap 95\% intervals (10{,}000 resamples of taps with
their claims): 0.385--0.506 for the RECAP model versus 0.040--0.085 for
the control. The parser emits up to three claims per explanation, but at
most two claims shared a tap here, so these barely widen the unclustered Wilson
intervals (0.384--0.506 / 0.042--0.089).
Table~\ref{tab:s4} gives the full readout behind the main text's numbers.

\begin{figure}[h]
\centering
\includegraphics[width=0.86\columnwidth]{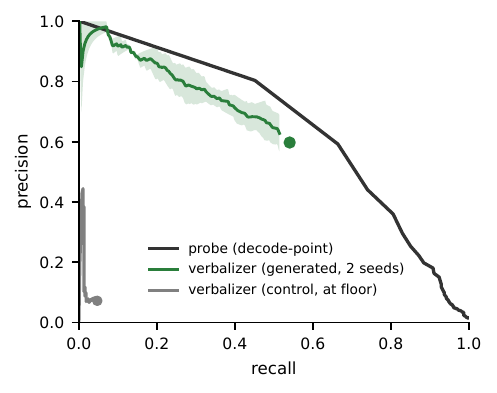}
\caption{Decodability exceeds verbalizability at scale (Pythia-160M, designated
content). A fresh probe reading the reconstructor's decode-point hidden state (black)
recovers the designated content across operating points; the reconstruction
verbalizer's \emph{generated} precision--recall curve (green, two reconstructor seeds)
lies below it, and its operating point sits near 0.60 precision. The content
is present and probe-decodable; free-form generation conveys only part of it. The
control verbalizer (grey) is at the floor.}
\label{fig:verbalizer_pr}
\end{figure}

\begin{table}[h]
\centering

\setlength{\tabcolsep}{4pt}
\begin{tabular}{lrr}
\toprule
metric (400 taps each) & control & RECAP \\
\midrule
stated word claims            & 406  & 252 \\
word truth                    & 0.062 & 0.444 \\
word grounded @ $\tau{=}0.02$ & 0.48 & 0.78 \\
word grounded @ $\tau{=}0.05$ & 0.29 & 0.65 \\
word grounded @ $\tau{=}0.10$ & 0.15 & 0.47 \\
grounded-minus-true gap       & $+0.42$ & $+0.33$ \\
token-at-$-8$ truth           & 0.043 & 0.039 \\
token-at-$-8$ grounded @ $\tau{=}0.05$ & 0.32 & 0.43 \\
mean $\rtilde$                & 0.19 & 0.24 \\
\bottomrule
\end{tabular}
\caption{Scale verbalizer audit, full readout (one seed per condition).
The control reproduces the grounded-and-false signature; the RECAP model
states words truly far more often (0.44 versus 0.06). The offset-8 truth and
grounding are computed over all taps, but the true offset-8 token is a
designated top-64 target in only 2.2\% of taps ($n{\approx}9$); its near-zero
truth is a base-rate artifact (the verbalizer is trained to emit a claim on
every tap), too few to characterize, so we report it as unrecovered rather
than as a code.}
\label{tab:s4}
\end{table}

\paragraph{Robustness to reconstructor strength.} Retraining the AV/AR pair
stronger (LoRA rank 32, more SFT, reconstructor, and rejection-sampling rounds)
on the same target models raises mean $\rtilde$ to 0.208 (control) and 0.300
(RECAP). The gap holds on both axes: RECAP states words truly at 0.574 versus
0.152 and, on the shared 381 present words, recovers them at recall 0.470
versus 0.037; the control states fewer words (92 versus 312) yet trails on
precision and recall alike, so its higher truth rate is selectivity. Offset-8
truth is again near zero (0.028/0.026), consistent with its 2.2\% storable-tap
base rate rather than a code. The result
is also stable across seeds: over the three class-balanced RECAP target-model
seeds the stated-word truth is 0.444, 0.455, 0.443, against three control NLA
seeds at 0.062, 0.060, 0.000.

\paragraph{Decodability exceeds verbalizability (readout redesign).} To locate
the gap between the verbalizer's truth and what the tap contains, we first
place the verbalizer on the fitted probe's precision--recall curve: at the
verbalizer's own recall the probe reaches ${\sim}0.80$ precision
(designated-content AUC 0.99), while the verbalizer generates at 0.44--0.63. The
fair comparison is the AUC-vs-precision mismatch, not a literal $0.99$-vs-$0.57$
gap. We then rebuild the reader: standardized (not per-sample
norm-stripped) injection, six injected tokens via an MLP adapter, MLP-LoRA, and
a decode-point presence head trained on the exact window-occurrence labels
(external truth, never a learned monitor). This makes the designated content
linearly readable from the reader's \emph{own} last-prompt hidden state at
$0.79$--$0.81$ (two seeds), yet its generated precision stays $0.56$--$0.63$; a
matched control reads $0.03$ at the same point under the identical head. A
per-word confidence sweep traces the verbalizer's generative PR curve entirely
below the probe's (Figure~\ref{fig:verbalizer_pr}). Reader scale
(0.5B vs.\ 3B), RAFT reconstruction reward, a direct truth reward, injection
bandwidth, and explicit negative supervision were each tried and none raises
generated precision, so the residual is consistent with a cost of serializing
sparse superposed features by generation rather than missing information.

\section{Hyperparameters and Reproducibility Notes}
\label{app:hyper}

\paragraph{Scale runs (Pythia-160M).} Continued pretraining on
pile-uncopyrighted, 500M tokens per run, identical data and order across
runs; batch $64\times1024$ tokens, AdamW ($10^{-4}$, cosine to 10\%, warmup
100, weight decay 0.01, gradient clip 1.0), bf16 autocast with fp32 weights.
Heads: linear on the layer-6 hidden state. Presence targets use BCE averaged
over $K$; the class-balanced variant weights positives by the empirical
inverse window base rate, clamped at 100, estimated from the first 2M
training tokens. Identity targets use cross-entropy over the top-$K$ content
vocabulary at offset $-8$. Fresh probes: closed-form ridge per rank bucket at fixed
$\lambda{=}10^4$, fit on the first 80\% of held-out probe sequences and evaluated
on the last 20\% (disjoint contiguous blocks; long documents may straddle the
boundary); identity probes are 64-class logistic. An earlier version selected
$\lambda$ per bucket from $\{10,10^2,10^3,10^4\}$ by evaluation-split AUC; a full
rerun with $\lambda$ fixed reproduces every reported bucket AUC to three decimals,
so no selection bias enters the reported numbers. Optimizer hyperparameters (learning rate, warmup, weight decay, gradient clip) were fixed to standard values and not swept; the method-specific hyperparameters (aux weight, target count~$K$, designation dose, tap layer, head capacity) are the ones varied, with ranges and the selection rule in the Ablation Index.
Held-out language-modeling loss is measured on 20M tokens. Significance is assessed throughout by 95\% bootstrap confidence intervals and disjoint-interval comparisons (with a Hanley--McNeil power analysis for the held-out-band null), not by fixed-level hypothesis tests. All experiments run
on single NVIDIA A100-80GB GPUs under Modal (Debian-slim, Python~3.12; PyTorch, Transformers, Accelerate, and NumPy, with exact versions pinned in the released environment file). One caution
uncovered during setup: recent library defaults load fp16 checkpoints
natively, and AdamW in fp16 destroys all weights in a single step through
epsilon underflow while loss and gradients still read finite; all runs force
fp32 parameters.

\paragraph{Sandbox runs.} Two synthetic scenarios (story; marketplace) over
closed vocabularies; target models are 8-layer, $d{=}256$ transformers
trained from scratch; the toy NLA uses a 0.5B-parameter AV/AR pair with a
dimension adapter, norm-matched activation injection, cold-start SFT on
full-story descriptions, then alternating reconstructor fitting and
rejection-sampling rounds against reconstruction reward (the released
recipe's structure). Audits parse stated specifics by exact closed-vocabulary
scan; grounding flips swap one stated word for a same-type vocabulary word
and measure the reconstruction drop. RECAP runs use aux weight 0.3, 3{,}000 steps, batch 128, learning rate $3{\times}10^{-4}$,
and 30{,}000 probe taps. Seeds: three (scenario 1) plus two (scenario 2); each run is controlled by a single integer seed passed to \texttt{torch.manual\_seed} (with a matched CUDA generator for sampling), fixing initialization, data order, and stochastic training and evaluation; RECAP
ceilings are evaluated by independently fitted probes on tuple-disjoint
splits in scenario 1 and held-out splits in scenario 2.

\end{document}